\pdfoutput=1

\documentclass[11pt]{article}

\usepackage[final]{ACL2023}

\usepackage{times}
\usepackage{latexsym}

\usepackage[T1]{fontenc}
\usepackage[utf8]{inputenc}

\usepackage{microtype}
\usepackage{inconsolata}
\usepackage{multirow} 
\usepackage{subcaption}  
\usepackage{graphicx}
\usepackage{booktabs}
\usepackage{xurl}
\usepackage{breakurl}

\title{No Language Data Left Behind: A Cross-Cultural Study of CJK Language Datasets in the Hugging Face Ecosystem}

\author{
Dasol Choi$^{1,2, 3}$\thanks{~Work done during internship at SIONIC AI.}\quad
Woomyoung Park$^{4}$\quad
Youngsook Song$^{5}$\thanks{~Corresponding author.}\\
$^{1}$Yonsei University \; $^{3}$AIM Intelligence \;  $^{2}$MODULABS \;
$^{4}$Sionic AI \; $^{5}$Lablup \\
\texttt{dasolchoi@yonsei.ac.kr} \quad \texttt{max@sionic.ai} \quad \texttt{yssong@lablup.com}
}

\begin{document}
\maketitle

\renewcommand{\thefootnote}{}

\begin{abstract}
Recent advances in Natural Language Processing (NLP) have underscored the crucial role of high-quality datasets in building large language models (LLMs). However, while extensive resources and analyses exist for English, the landscape for East Asian languages, particularly Chinese, Japanese, and Korean (CJK), remains fragmented and underexplored, despite these languages serving over 1.6 billion speakers. To address this gap, we investigate the HuggingFace ecosystem from a cross-linguistic perspective, focusing on how cultural norms, research environments, and institutional practices shape dataset availability and quality. Drawing on more than 3,300 datasets, we employ quantitative and qualitative methods to examine how these factors drive distinct creation and curation patterns across Chinese, Japanese, and Korean NLP communities. Our findings highlight the large-scale and often institution-driven nature of Chinese datasets, grassroots community-led development in Korean NLP, and an entertainment and subculture-focused emphasis on Japanese collections. By uncovering these patterns, we reveal practical strategies for enhancing dataset documentation, licensing clarity, and cross-lingual resource sharing, guiding more effective and culturally attuned LLM development in East Asia. We conclude by discussing best practices for future dataset curation and collaboration, aiming to strengthen resource development across all three languages.
\end{abstract}

\section{Introduction}

With the emergence of Large Language Models (LLMs) transforming the field of Natural Language Processing (NLP)\cite{bert, gpt3, gpt4}, the importance of high-quality datasets in model development has become increasingly critical. For datasets to be valuable in this context, they must meet both quantitative requirements (sufficient size and coverage) and qualitative standards (reliability and representativeness). While English-language resources have been extensively studied, the landscape of datasets for East Asian languages—particularly Chinese, Japanese, and Korean (CJK)—remains comparatively underexplored\cite{joshi2020state, bender2019benderrule}. This gap is especially noteworthy given that these languages collectively serve over 1.6 billion speakers and originate from major hubs of technological innovation.

\sloppy In recent years, platforms such as HuggingFace have emerged as central repositories for distributing and accessing NLP datasets, making these resources widely accessible while introducing new challenges in dataset discovery, quality assessment, and cross-lingual collaboration\cite{huggingface, lhoest2021datasets}. These challenges are particularly pronounced for CJK languages due to their unique linguistic features, distinct cultural contexts, and varying approaches to data sharing and documentation.

While CJK languages play an increasingly important role in global NLP research, several critical issues need to be addressed. First, there is a limited understanding of how dataset creation patterns differ across these language communities—and how those differences reflect their respective NLP ecosystems. Second, although cultural and institutional factors evidently influence dataset generation, their specific impacts on dataset characteristics and quality have yet to be systematically investigated. Third, the potential for cross-lingual synergies among CJK languages remains largely untapped, despite their many shared cultural and linguistic foundations. For instance, while benchmarks like MMLU\cite{mmlu} are being generated in multiple languages, insufficient comparative analysis exists regarding how these variants differ from traditional parallel corpora or how download patterns across regions reflect distinct cultural preferences.
To address these challenges, this study investigates the HuggingFace ecosystem for Chinese, Japanese, and Korean NLP resources, focusing on dataset development and usage patterns. The specific objectives of this work are to:

\begin{itemize}
   \item Examine the current landscape of CJK datasets on HuggingFace, including key metadata such as domain, dataset size, documentation practices, and usage statistics.
   \item Analyze the cultural characteristics and development patterns underlying dataset creation in each language community, highlighting commonalities and differences in curation and documentation.
   \item Identify potential synergies and cross-lingual opportunities among the three languages, and propose strategies for more effective, collaborative dataset development.
\end{itemize}

\section{Related Work}

Documentation efforts for CJK language resources have expanded over time. For Chinese, \citet{tao2009advances} introduce systematic approaches for constructing and evaluating linguistic data resources, while \citet{li2023chinese} compile a comprehensive repository emphasizing accessibility and classification frameworks. In the Korean context, \citet{cho2020open} survey existing NLP resources, and \citet{cho2023revisiting} examine how local research cultures influence resource development. However, these studies largely predate large generative models and rely on GitHub as the primary hub, while systematic analyses of CJK datasets within newer ecosystems like HuggingFace remain rare.

Major multilingual projects including BigScience\citep{le2023bloom}, CC100 \citep{wenzek2019ccnet}, and LAION \citep{schuhmann2022laion} expand non-English data availability, yet don't investigate cultural factors affecting dataset usage within communities. The movement toward improved dataset documentation \citep{gebru2021datasheets} highlights ethical considerations, though adoption varies across languages and platforms.

HuggingFace has emerged as a central repository for NLP resources, but studies flag challenges including inconsistent documentation \citep{yang2024navigating}, limited transparency \citep{pepe2024hugging}, and ambiguous licensing. These issues underscore the need to examine dataset curation in cultural contexts \citep{lhoest2021datasets}. Recent work by \citet{dargis2024evaluating} offers insights for building evaluation frameworks for languages with particular traits, yet no study comprehensively analyzes how CJK dataset ecosystems are shaped by local research cultures, licensing preferences, and community-driven development.

Building on these perspectives, our research provides the first large-scale, comparative analysis of CJK datasets within the HuggingFace ecosystem, investigating how cultural contexts and documentation standards influence dataset usage patterns and resource quality.

\begin{table*}[h]
\centering
\renewcommand{\arraystretch}{1.1}
\small
\begin{tabular}{lp{11.5cm}}
\hline
\textbf{Category} & \textbf{Evaluation Metrics} \\
\hline
\multirow{3}{*}{Scale \& Composition} 
  & \textbf{Dataset Size:} Distribution across size categories (small, medium, large, extra-large) \\
  & \textbf{Language Makeup:} Distribution of monolingual, English-paired, and multilingual datasets \\
  & \textbf{Task Types:} Distribution of major NLP tasks (text generation, QA, classification, etc.) \\
\hline
\multirow{3}{*}{Development Patterns} 
  & \textbf{Ownership Structure:} Proportions of corporate, institutional, and individual contributions \\
  & \textbf{License Types:} Distribution of permissive, copyleft, unknown, and other licenses \\
  & \textbf{Community Activity:} Dataset creation trends and instruction tuning development \\
\hline
\multirow{3}{*}{Documentation Quality} 
  & \textbf{Academic Validation:} Presence of associated arXiv papers and research citations \\
  & \textbf{Documentation Standards:} Adherence to HuggingFace dataset card templates \\
  & \textbf{Documentation Depth:} Comprehensiveness of dataset descriptions and README files \\
\hline
\multirow{3}{*}{Cultural Characteristics} 
  & \textbf{Domain Focus:} Specialized fields (e.g., medical, entertainment, content moderation) \\
  & \textbf{Resource Development:} Approaches to dataset creation and curation \\
  & \textbf{Community Priorities:} Language-specific preferences and development patterns \\
\hline
\end{tabular}
\caption{Analysis framework for CJK datasets, organized by category.}
\label{tab:analysis_framework}
\end{table*}

\begin{figure}[h!]
    \includegraphics[width=0.48\textwidth]{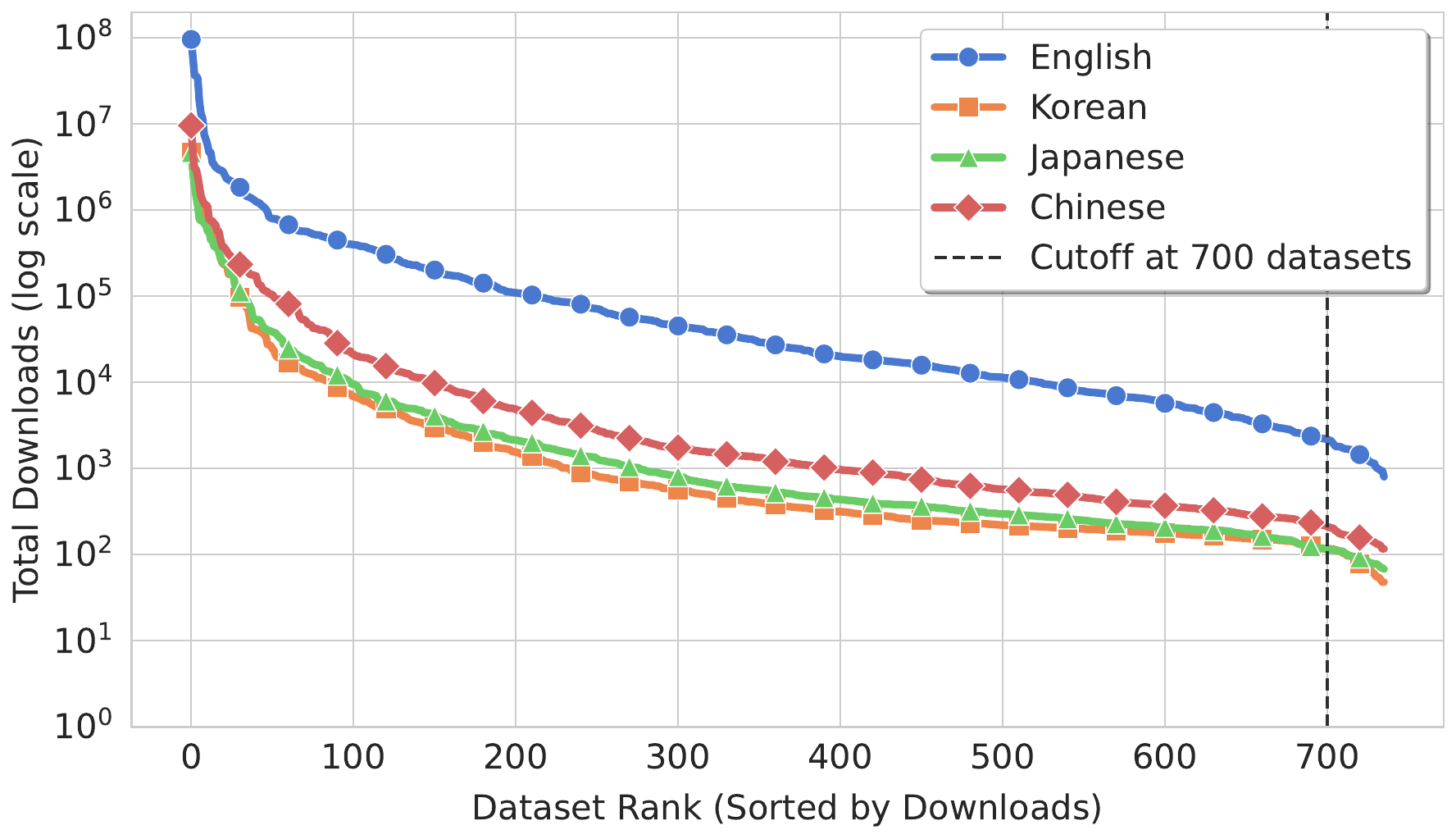}
    \caption{Datasets of each language sorted by number of downloads in descending order. Based on the decreasing pattern of downloads, we set the cutoff point at 700.}
    \label{fig:download-stats}
\end{figure}

\section{Method}

\subsection{Data Collection}

We systematically collected dataset information from the HuggingFace platform using their Datasets API\footnote{Datasets were retrieved from \url{https://huggingface.co/datasets} using language filters and download frequency sorting. All statistics were recorded on January 28, 2025.}. Our data collection strategy focused on identifying actively used datasets for each target language (Chinese, Japanese, Korean, and English as a reference).

To determine a cutoff point for dataset inclusion, we analyzed the download frequency distribution for each language (Figure~\ref{fig:download-stats}). The download counts follow a power-law distribution, with all languages showing consistent patterns. The distributions converge around the 700th dataset, where download counts fall below 100. Beyond this point, we observe minimal engagement and declining documentation quality. This natural boundary led us to set our cutoff at 700 datasets per language, ensuring both coverage and quality.

For each dataset, we extracted metadata across four categories: \textbf{Scale \& Composition} (dataset size, language combinations, task types, temporal patterns), \textbf{Development Patterns} (ownership, licensing, community metrics), \textbf{Documentation Quality} (dataset cards, citations, README files, metadata completeness), and \textbf{Cultural Characteristics} (domain focus, development approaches, community patterns).

In addition, we collected the complete dataset cards to analyze documentation practices and cultural characteristics in depth. We will release the full metadata and dataset card contents as a public resource upon publication.


\subsection{Analysis Framework}

Our analysis framework combines quantitative and qualitative approaches to examine CJK language datasets. Table~\ref{tab:analysis_framework} presents our analysis metrics across four main categories. For quantitative analysis, we focus on measuring dataset sizes, language distributions, task type proportions, ownership ratios, and documentation completeness. Our qualitative analysis examines domain preferences, resource development approaches, and community characteristics. This mixed-method approach helps us understand how dataset development patterns reflect each language community's unique characteristics, particularly in terms of instruction tuning trends, domain preferences, and resource development strategies.

\begin{figure*}[t]
    \centering
    \begin{subfigure}[b]{0.37\textwidth}
        \includegraphics[width=\textwidth]{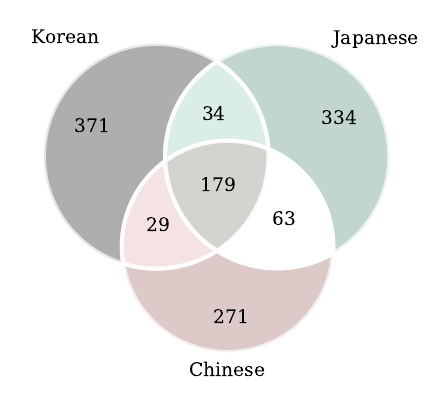}
        \caption{Distribution of CJK Language Intersection}
        \label{fig:language-venn}
    \end{subfigure}
    \quad  
    \begin{subfigure}[b]{0.5\textwidth}
        \hspace*{0.5cm}
        \includegraphics[width=\textwidth]{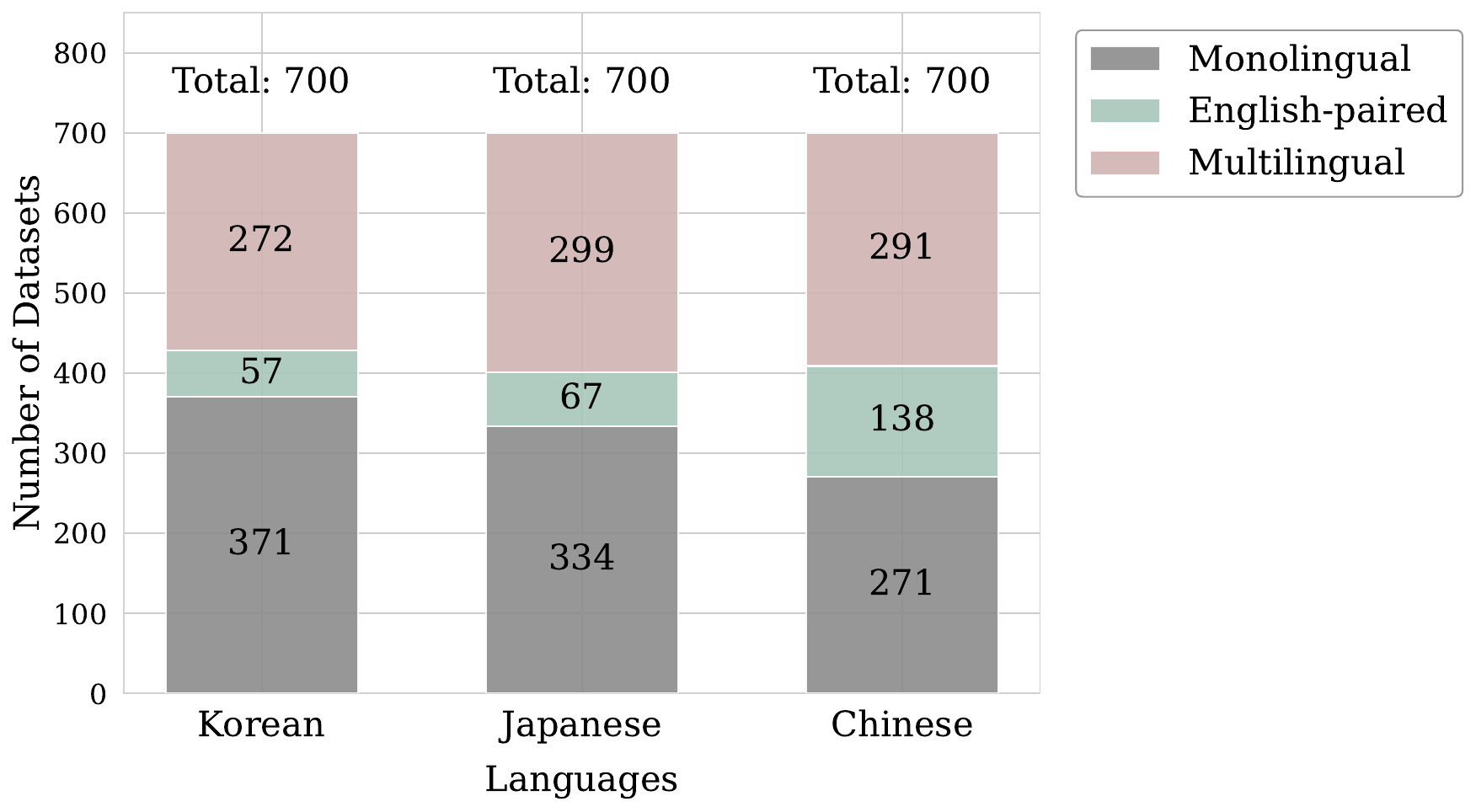}
        \caption{Composition of Language Types in Datasets}
        \label{fig:language-bar}
    \end{subfigure}
    \caption{Distribution and Composition Analysis of CJK Language Datasets. (a) Illustrates the intersections among CJK language datasets, showing unique and overlapping dataset counts. (b) Shows the composition of the top 700 downloaded datasets for each language, categorized into monolingual, English-paired, and multilingual resources.}
    \label{fig:language-overview}
\end{figure*}

\section{Results and Analysis}

\subsection{Overview of CJK Datasets}

The stacked bar chart (Figure~\ref{fig:language-bar}) provides a view of the top 700 most downloaded datasets for each language, categorizing them into monolingual, English-paired (bilingual with English), and multilingual (three or more languages) resources. In this broader analysis, Chinese datasets show the highest proportion of English-paired resources (148 datasets), notably higher than Korean (57) or Japanese (67), suggesting a greater emphasis on cross-lingual applications. The multilingual category shows substantial representation across all three languages, with similar proportions (Korean: 272, Japanese: 299, Chinese: 291), indicating active participation in multilingual resource development. While multilingual resources reflect diverse aspects of CJK datasets, we focus our subsequent analyses on \emph{monolingual datasets} to better understand language-specific characteristics.

\paragraph{Dataset Size Distribution.}

Table~\ref{tab:size-distribution} presents the size distribution of monolingual datasets across languages, categorized as Small (<10MB), Medium (10MB–100MB), Large (100MB–1GB), and Extra-Large (>1GB). Chinese datasets show a relatively balanced distribution across size categories, with a notable presence in large (10) and extra-large (12) categories. Japanese datasets demonstrate strong representation in small-scale resources (123) but notably lack extra-large datasets. Korean shows similar concentration in small datasets (137) and limited presence in medium and large categories, yet maintains a notable presence in the extra-large category (7). English datasets maintain the highest counts across all categories, providing a reference point for resource availability.

\begin{table}[h]
\centering
\small  
\renewcommand{\arraystretch}{1.1}
\setlength{\tabcolsep}{4pt}  
\begin{tabular}{lcccc}
\toprule
Size & English & Chinese & Japanese & Korean \\
\midrule
\textbf{S} (<10M)   & 258  & 144  & 123  & 137  \\
\textbf{M} (10M–100M)  & 21   & 11   & 3    & 3    \\
\textbf{L} (100M–1B)   & 14   & 10   & 4    & 3    \\
\textbf{XL} (>1B)   & 22   & 12   & -    & 7    \\
\bottomrule
\end{tabular}
\caption{Dataset size distribution across languages.}
\label{tab:size-distribution}
\end{table}

\begin{figure}[h!]
    \includegraphics[width=0.48\textwidth]{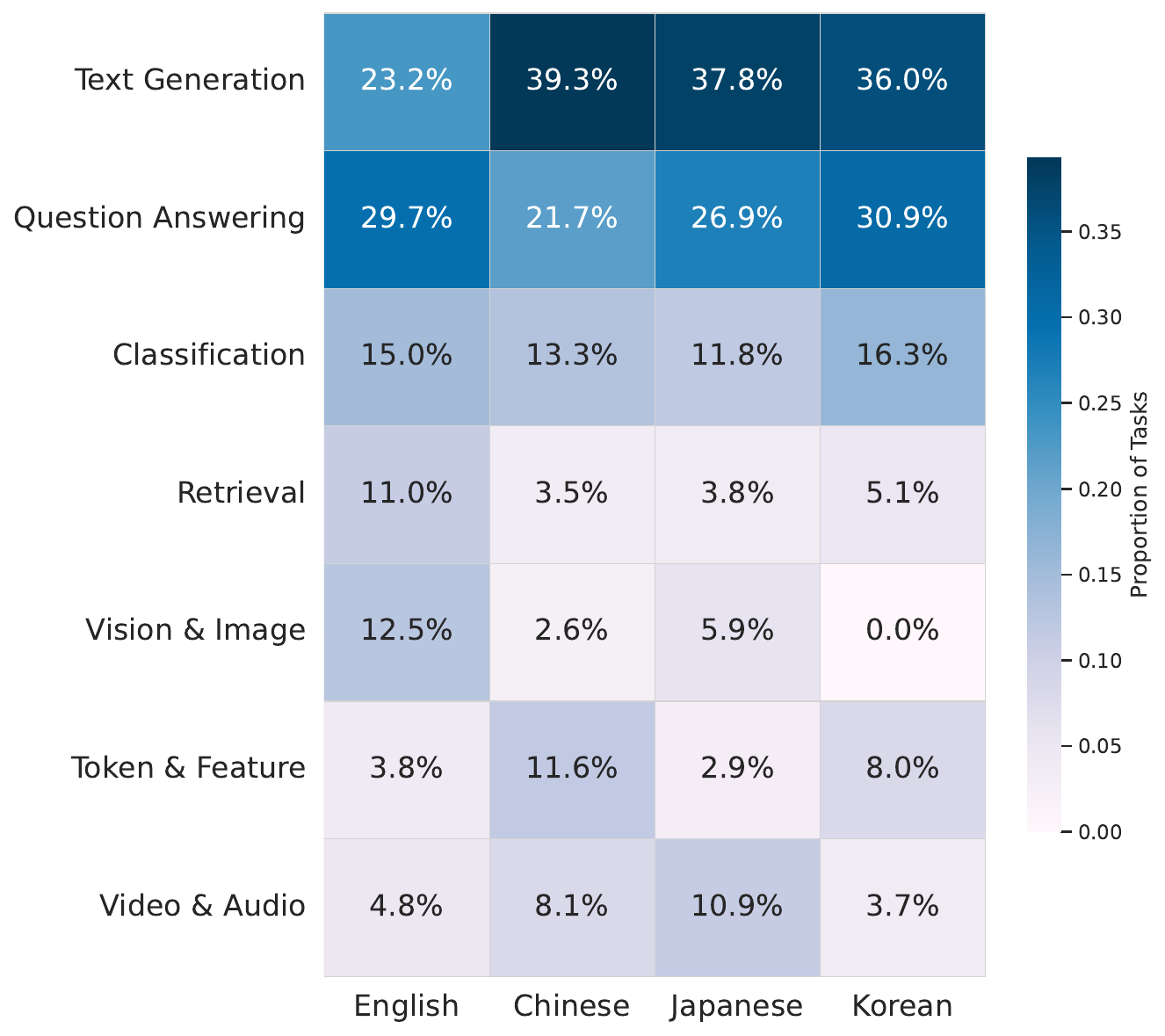}
    \caption{Task distribution across different languages. The heatmap illustrates the proportion of datasets belonging to the top 7 most frequent task categories across English, Chinese, Japanese, and Korean datasets.}
    \label{fig:task-distribution}
\end{figure}

\subsection{Comparative Characteristics of CJK Datasets}

\paragraph{Task Distribution}

Figure~\ref{fig:task-distribution} presents the distribution of task categories across languages through a heatmap visualization\footnote{Task categories group related NLP tasks based on their functionality (e.g., Text Generation includes text-generation, language-modeling, fill-mask; Question Answering includes question-answering, multiple-choice, extractive-qa).}. \emph{Text Generation} emerges as the dominant task across all languages, with particularly high proportions in Chinese (39.3\%), Japanese (37.8\%), and Korean (36.0\%) datasets compared to English (23.2\%). \emph{Question Answering} follows as the second most common task, with moderate variations: Korean (30.9\%) and English (29.7\%) show higher proportions than Chinese (21.7\%) and Japanese (26.9\%).

The analysis reveals distinctive task preferences across languages beyond these two categories. \emph{Classification} tasks appear between 11.8\% and 16.3\% of datasets, with Korean (16.3\%) having the highest ratio and Japanese (11.8\%) the lowest. \emph{Token \& Feature} tasks are more prominent in Chinese (11.6\%) and Korean (8.0\%) than in English (3.8\%) or Japanese (2.9\%). \emph{Video \& Audio} tasks show varied representation, with Japanese leading at 10.9\%, followed by Chinese (8.1\%), English (4.8\%), and Korean (3.7\%). Lastly, \emph{Vision \& Image} tasks exhibit particularly striking differences: English leads at 12.5\%, followed by Japanese (5.9\%) and Chinese (2.6\%), while Korean shows no representation (0.0\%). \emph{Retrieval} tasks also show notable variation, with English (11.0\%) significantly ahead of other languages (3.5-5.1\%).

These distinctive patterns reflect differing research priorities and technological needs across language communities. The Japanese emphasis on Video \& Audio may correspond to its strong anime and entertainment industry, while Korean's focus on Classification and Chinese's on Token \& Feature tasks suggest prioritization of fundamental NLP infrastructure development tailored to their respective linguistic complexities.

\begin{figure*}[h!]
    \centering    \includegraphics[width=1\linewidth]{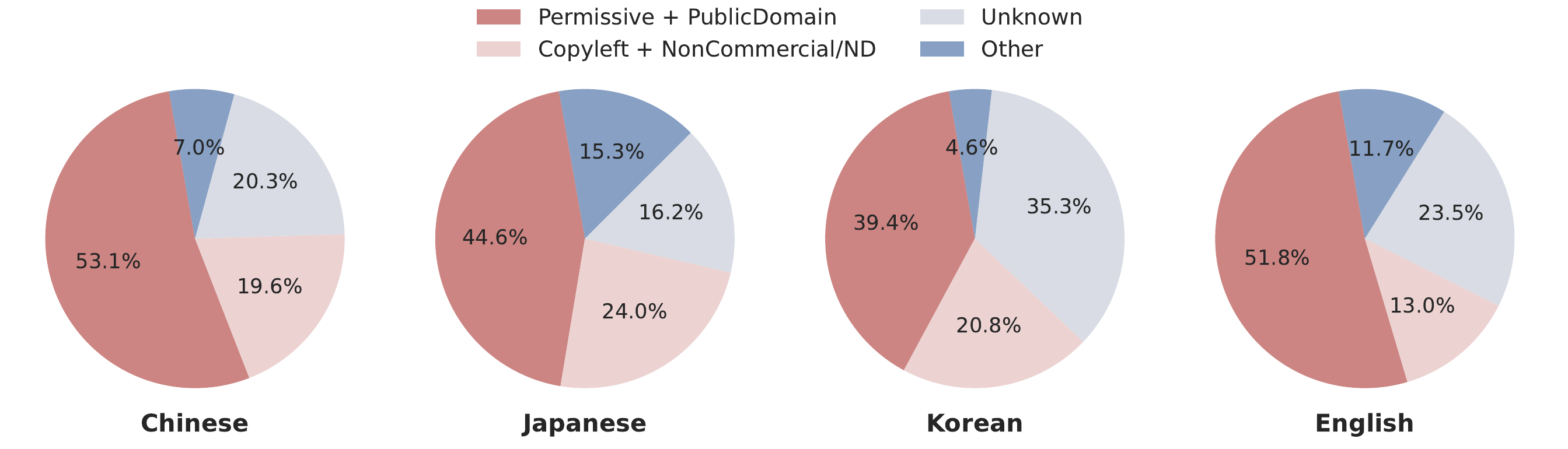}
    \caption{License distribution across CJK and English datasets, showing the proportion of \emph{Permissive + PublicDomain}, \emph{Copyleft + NonCommercial/ND}, 
\emph{Unknown}, and \emph{Other} licenses for each language community.}
    \label{fig:license_dist}
\end{figure*}

\begin{figure}[h!]
    \centering
    \hspace{-0.02\textwidth}
    \includegraphics[width=1\linewidth]{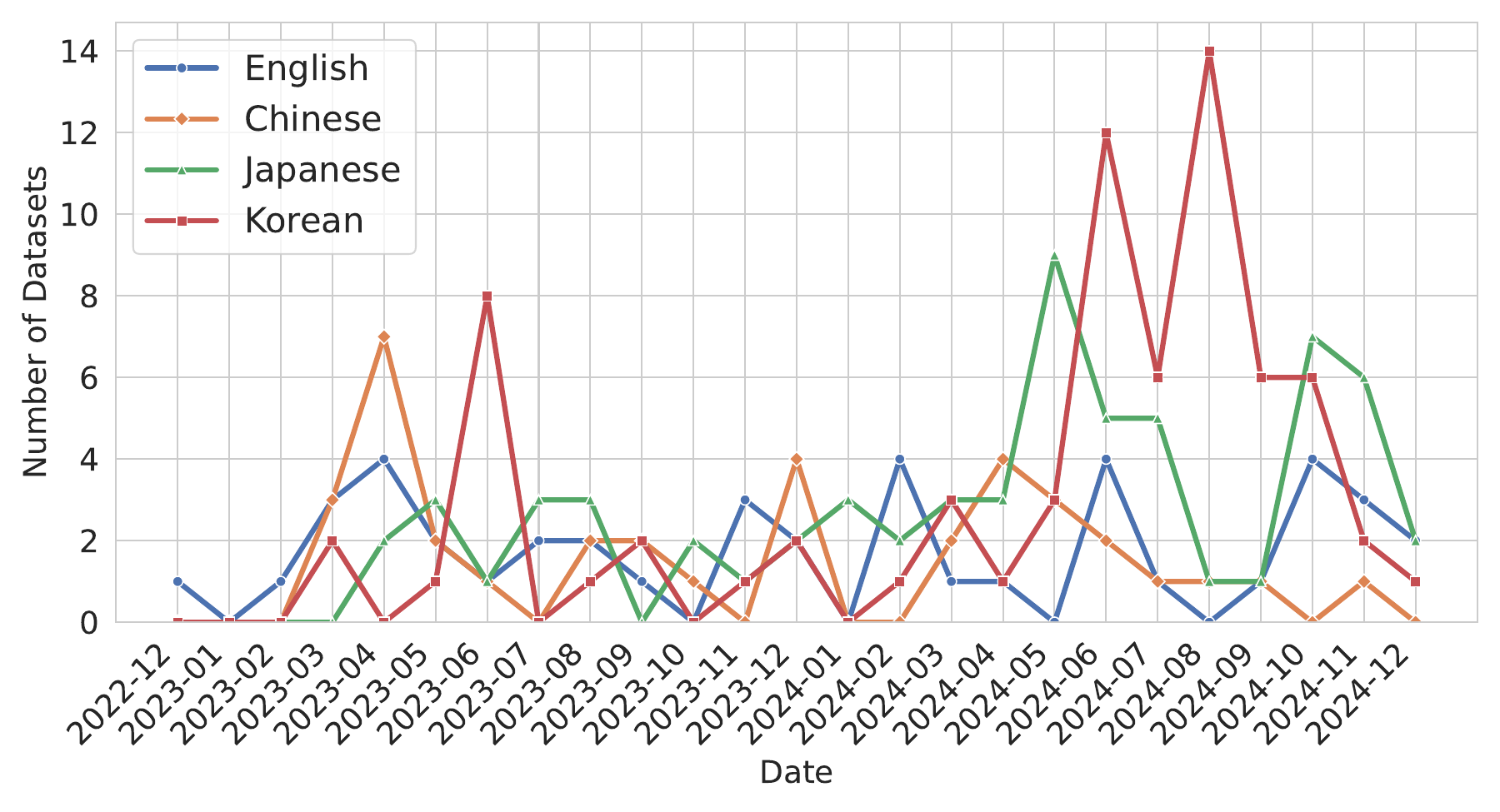}
    \caption{Instruction Datasets Over Time by Language (English and CJK), from late 2022 to 2024.}
    \label{fig:instruction_trends}
\end{figure}

\paragraph{Evolution of Instruction Tuning}

Analysis of instruction-tuning datasets\footnote{We identify instruction datasets through `instruct' keywords and common dataset names in metadata.} reveals distinct patterns across languages. Among the top 700 most downloaded datasets for each language, instruction-tuning datasets show notable presence: Korean (13.3\%), Japanese (12.6\%), Chinese (12.3\%), and English (7.0\%). This higher proportion in CJK languages compared to English suggests particularly active instruction-tuning development in these communities.

Analysis of temporal patterns (Figure~\ref{fig:instruction_trends}) reveals distinct characteristics across languages. Chinese datasets show notable early activity in 2023, peaking around 7 releases. Korean datasets demonstrate dramatic fluctuations in 2024, reaching highest peaks of 12-14 releases mid-2024. Japanese datasets show moderate initial activity but increased activity during 2024, reaching peaks of 7-9 releases. English datasets maintain stable patterns throughout, typically with 1-4 releases monthly.

These patterns reflect different community approaches to instruction dataset development:

\begin{itemize}
\item \textbf{Chinese:} Early adoption with moderate peaks (around 7 releases) followed by decreased activity
\item \textbf{Korean:} Shows the highest peaks (up to 14 releases) with considerable volatility
\item \textbf{Japanese:} Late but substantial increase in development activity
\item \textbf{English:} Consistent but moderate release patterns throughout
\end{itemize}

\begin{figure*}[h!]
    \centering
    \begin{subfigure}[b]{0.31\linewidth}
        \includegraphics[width=\linewidth]{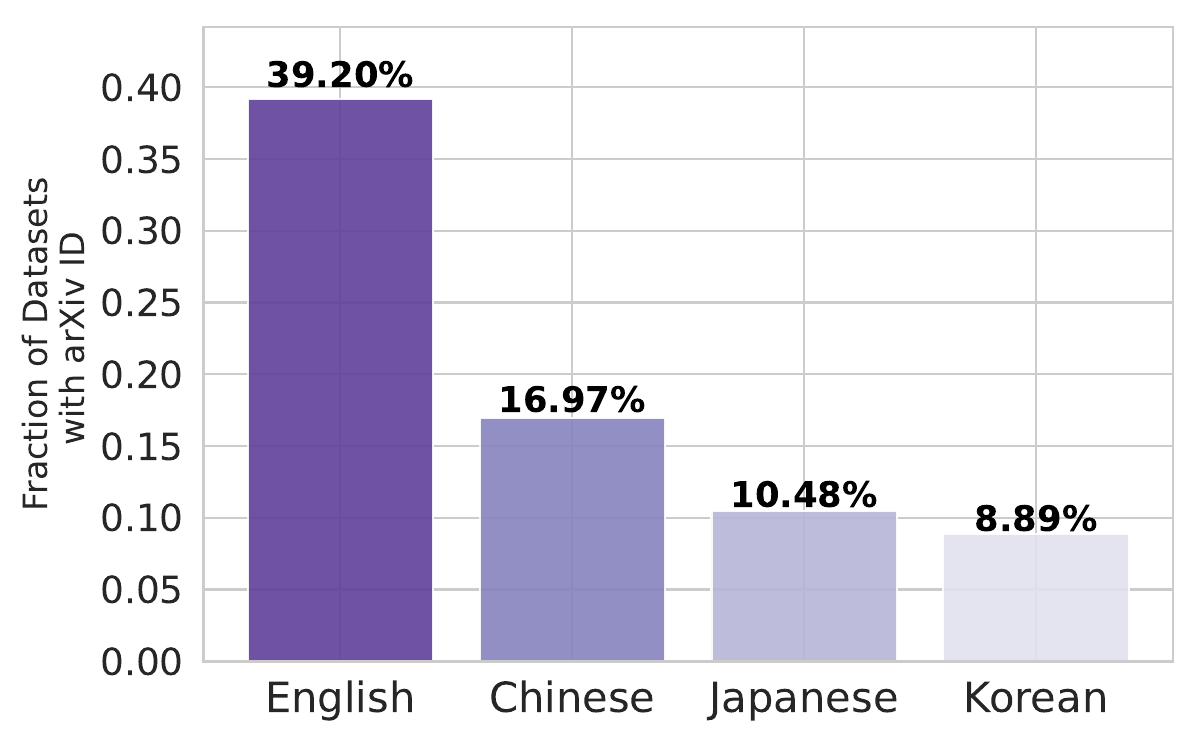}
        \caption{arXiv Paper Association}
        \label{fig:arxiv}
    \end{subfigure}
    \hspace{0.01\textwidth}
    \begin{subfigure}[b]{0.31\linewidth}
        \includegraphics[width=\linewidth]{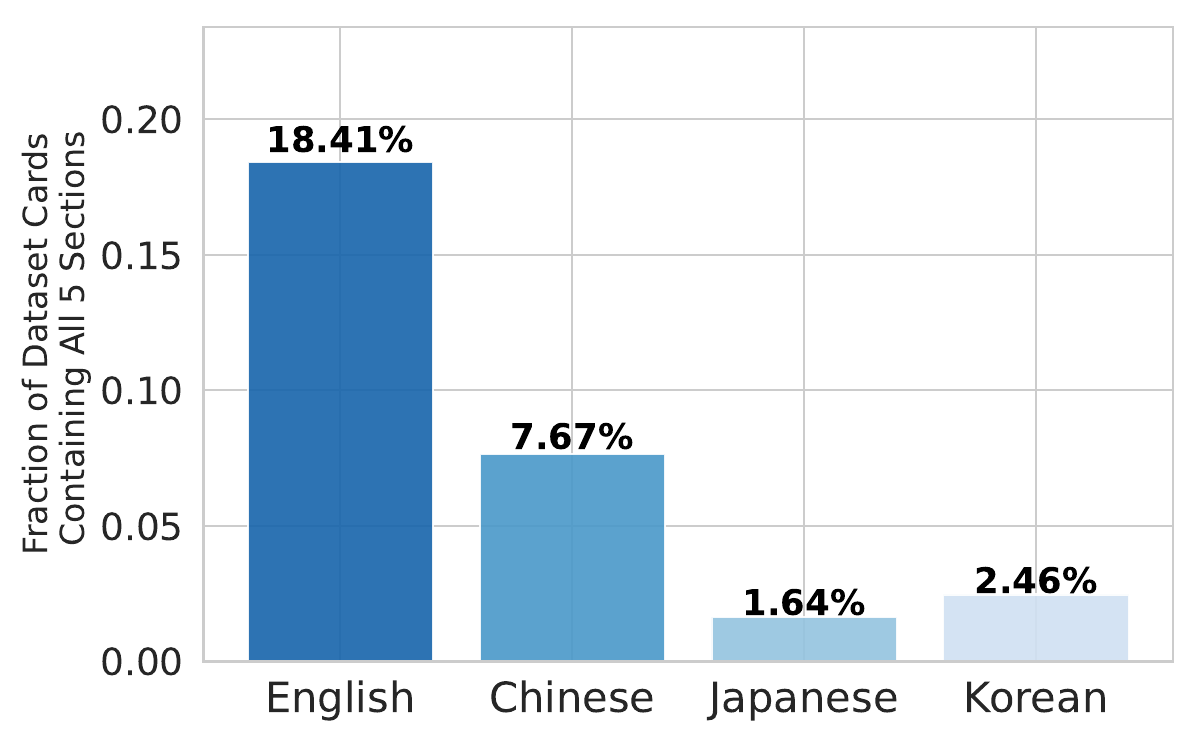}
        \caption{Template Adherence}
        \label{fig:template}
    \end{subfigure}
    \hspace{0.01\textwidth}
    \begin{subfigure}[b]{0.31\linewidth}
        \includegraphics[width=\linewidth]{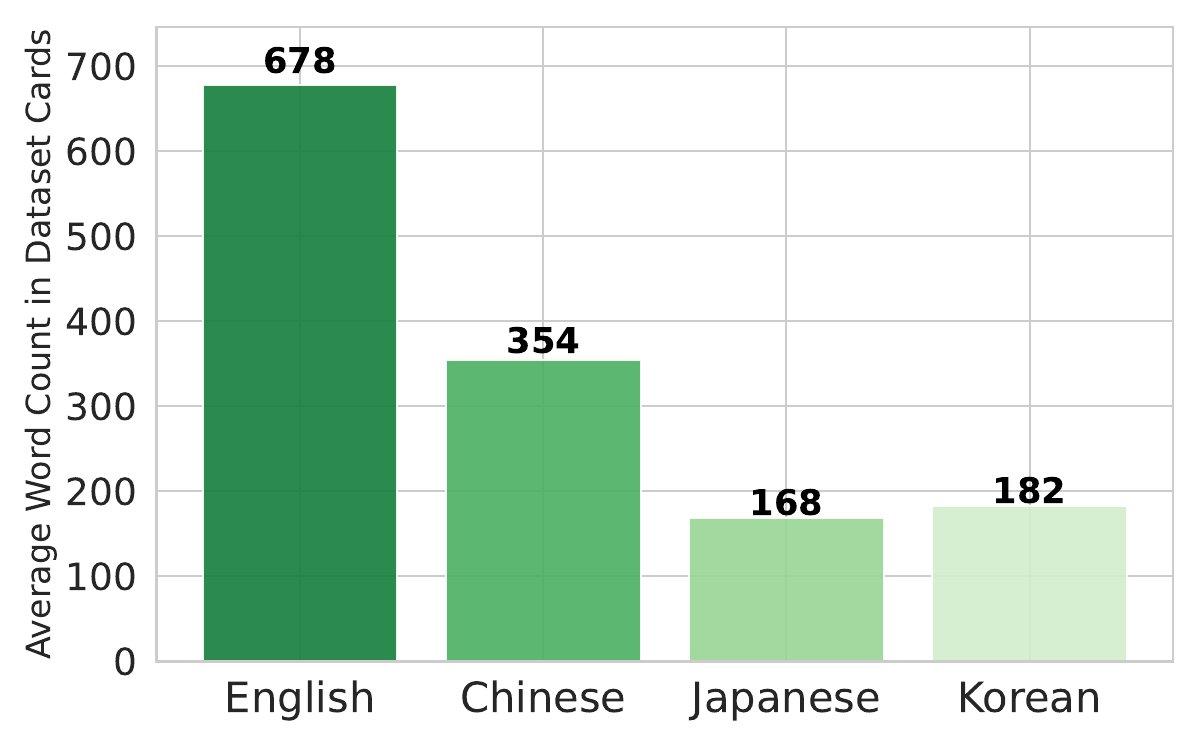}
        \caption{Documentation Length}
        \label{fig:wordcount}
    \end{subfigure}
    \caption{Comparison of Documentation Quality Across Languages. (a) Shows the percentage of datasets associated with academic publications, (b) presents the percentage of dataset cards containing all five structured sections, and (c) displays the average word count in dataset documentation.}
    \label{fig:documentation_metrics}
\end{figure*}

\paragraph{License Distribution Patterns}
Figure~\ref{fig:license_dist} shows how each language community approaches data licensing\footnote{Licenses were classified into five categories: Permissive (e.g., Apache, MIT, CC-BY), Public Domain (e.g., CC0, PDDL), Copyleft/ShareAlike (e.g., GPL, CC-BY-SA), NonCommercial/ND (CC-BY-NC, CC-BY-ND), and Other. For visualization, Permissive and Public Domain categories were combined, as were Copyleft/ShareAlike and NonCommercial/ND.}. Two major observations emerge. First, Chinese (53.1\%) and English (51.8\%) exhibit the highest proportions of \emph{Permissive} or \emph{Public Domain} licenses, indicating a shared culture of open access. Both also include moderate segments of \emph{Unknown} (Chinese 20.3\%, English 23.5\%) and \emph{Copyleft/NonCommercial} (19.6\% and 13.0\% respectively), suggesting a balance between openness and controlled usage.

Second, Japanese (44.6\%) and Korean (39.4\%) have lower shares of \emph{Permissive/Public Domain} compared to Chinese and English but differ substantially in other categories. Japanese devotes 24.0\% to \emph{Copyleft/NonCommercial}—more than any other language—while also having a relatively high \emph{Other} category (15.3\%). In contrast, Korean stands out for its large \emph{Unknown} portion (35.3\%), underscoring possible gaps in documentation practices despite still having a notable \emph{Copyleft/NonCommercial} share (20.8\%). Taken together, these variations reflect distinct cultural, institutional, and legal factors influencing dataset license norms across CJK and English communities.

\begin{figure*}[h!]
    \centering
    \includegraphics[width=0.84\linewidth]{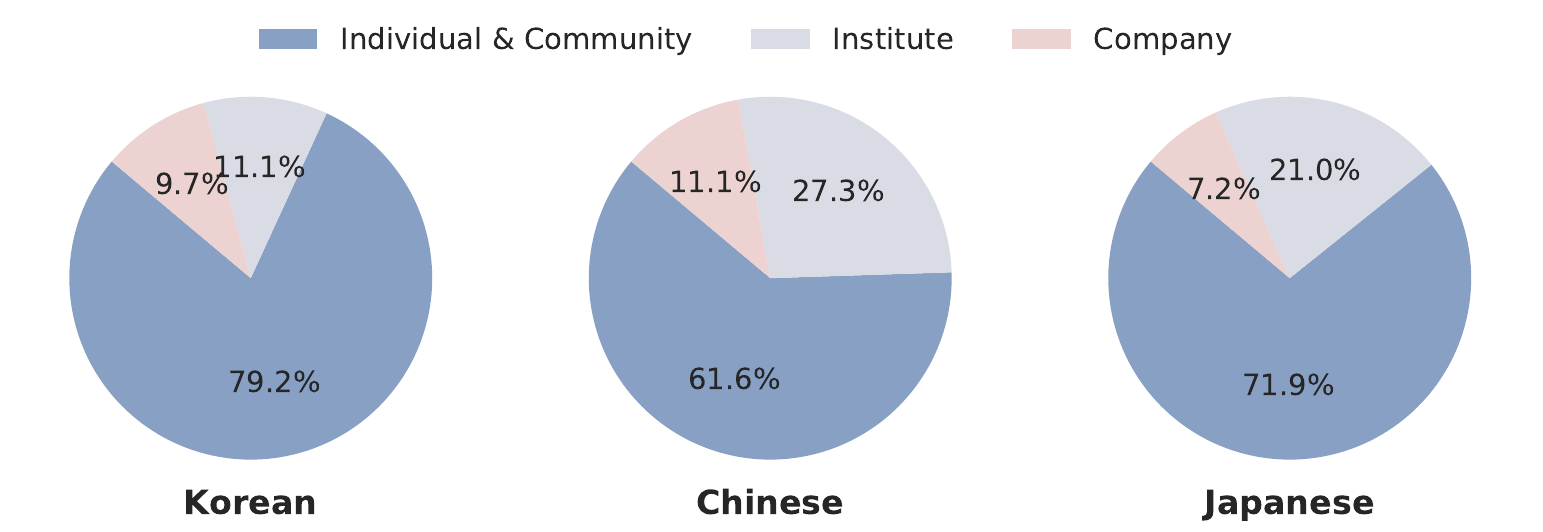}
    \caption{Dataset ownership across Korean, Chinese, and Japanese datasets, 
    illustrating the proportions of individual and community contributors, research institutes, and companies.}
    \label{fig:ownership_dist}
\end{figure*}

\paragraph{Documentation Quality Patterns}

Figure~\ref{fig:documentation_metrics} illustrates three key metrics related to dataset documentation. For academic grounding, measured by the presence of associated arXiv papers, English datasets lead with 39.20\%, followed by Chinese (16.97\%), while Japanese (10.48\%) and Korean (8.89\%) trail behind. Regarding structural completeness, measured by the presence of all five recommended Hugging Face dataset card sections, English again leads (18.41\%), with Chinese (7.67\%), Japanese (1.64\%), and Korean (2.46\%) showing lower completeness \footnote{HuggingFace dataset card sections: \emph{Description}, \emph{Structure}, \emph{Creation}, \emph{Usage}, and \emph{Additional Info}}. In terms of documentation depth, measured by average word count, English maintains the highest average (678 words), Chinese stands at 354, and Japanese and Korean remain lower at 168 and 182 respectively. Taken together, these figures indicate a consistent trend: English datasets demonstrate the most thorough and standardized documentation practices, Chinese resources show moderate completeness and Japanese and Korean documentation remains comparatively succinct or under-documented.

\paragraph{Dataset Ownership Patterns}

Figure \ref{fig:ownership_dist} illustrates the proportions of datasets contributed by individual and community contributors, research institutes, and companies in the Korean, Chinese, and Japanese communities\footnote{Analysis limited to CJK datasets with verifiable institutional affiliations.}. Individual and community contributors dominate across all three languages: Korean datasets lead with 79.2\%, followed by Japanese at 71.9\% and Chinese at 61.6\%. Research institutes play a stronger role in China (27.3\%) than in Korea (11.1\%) or Japan (21.0\%), suggesting more prominent institutional involvement in Chinese NLP resource development. Company contributions remain the smallest category across all three, though slightly higher in Chinese (11.1\%) compared to Korean (9.7\%) and Japanese (7.2\%). This highlights the key role of grassroots efforts as primary drivers of dataset creation.

\subsection{Language-Specific Characteristics}

\subsubsection{Chinese Dataset Ecosystem}

\paragraph{Comprehensive Evaluation Frameworks}
Chinese NLP resources on Hugging Face frequently feature large-scale, well-structured evaluation suites. For example, \path{ceval/ceval-exam} \cite{cevl} provides 13{,}948 multiple-choice questions across 52 domains, and \path{haonan-li/cmmlu} \cite{cmmlu} covers 67 subject areas spanning elementary to advanced professional levels. These broad assessments facilitate detailed benchmarking of model performance across diverse knowledge domains.

\paragraph{Specialized Medical Domain Resources}
Chinese datasets also demonstrate significant depth in specialized fields. The \path{FreedomIntelligence/CMB} \cite{cmedbenchmark} collection features a hierarchical structure (6 main categories, 28 subcategories) spanning 11{,}200 medical questions, thus enabling targeted evaluations in clinical question-answering. Similarly, \path{TCMLM/TCM_Humanities} \cite{TCM_Humanities} focuses on Traditional Chinese Medicine, integrating resources from professional certification materials and historical texts.

\paragraph{Dialectal and Cultural Diversity}
In addition to specialized domains, Chinese datasets often emphasize linguistic diversity and cultural preservation. The \path{Nexdata/chinese_dialect} \cite{nexdata_chinese_dialect} corpus contains 25{,}000 hours of dialect speech data, facilitating fine-grained dialect modeling. Likewise, \path{raptorkwok/cantonese-traditional-chinese-parallel-corpus} \cite{raptorkwok_cantonese} offers over 130k aligned sentence pairs for Cantonese--Mandarin translation. Future efforts could enrich such dialectal resources by detailing speaker demographics, annotation workflows, and language-specific quirks—thereby promoting more equitable research coverage across China's diverse linguistic communities.

\subsubsection{Korean Dataset Ecosystem}

\paragraph{Community-Driven Development and Its Impact}
Korean datasets on Hugging Face frequently emerge from grassroots, community-led efforts, rather than purely institutional or corporate projects. Prominent examples include contributions from open-source communities like \path{HAERAE}, which developed the widely-used \path{HAERAE-HUB/KMMLU} benchmark~\citep{kmmlu}, as well as individual developers such as \path{beomi}, \path{maywell}, and \path{taeminlee}, who have created highly-downloaded resources. Even widely-used benchmarks like \path{klue/klue}~\citep{klue} represent collaborative efforts between academia, industry, and individual researchers rather than single-entity projects. The broader ecosystem is dominated by individual and community contributors who account for 79.2\% of Korean datasets. This community-driven approach has accelerated the proliferation of new resources but also contributed to inconsistencies in documentation and licensing. For instance, Korean has the highest proportion of ``Unknown'' licenses among CJK languages (35.3\%), indicating gaps in legal clarity and potential challenges for commercial or cross-institutional usage. Moreover, only 8.89\% of Korean datasets are linked to an arXiv publication—lower than both Chinese (16.97\%) and Japanese (10.48\%). These factors may hinder collaborative research or reproducibility, underscoring the need for more standardized dataset cards \cite{gebru2021datasheets} and explicit licensing.

\paragraph{Content Moderation Focus}

A unique strength of the Korean dataset ecosystem is its emphasis on content moderation, encompassing hate-speech detection, toxicity filtering, and profanity masking. Popular resources such as \path{jeanlee/kmhas-korean-hate-speech} \cite{kmhas}, \path{humane-lab/K-HATERS} \cite{khaters} and \path{Dasool/KoMultiText} \cite{komultitext} reflect heightened community and research interest in combating harmful or discriminatory language online. However, these moderation-oriented resources raise broader ethical and regulatory questions, such as defining thresholds for \emph{hate speech} or handling user privacy. Although the Hugging Face platform provides general community guidelines, more detailed policies—particularly for age-restricted or sensitive data—would help standardize safe usage of these potentially sensitive resources.\footnote{As of January 2025, based on Hugging Face’s publicly available platform policies and community guidelines.}

\subsubsection{Japanese Dataset Ecosystem}

\paragraph{Strong focus on subcultural content}
Japanese NLP datasets often integrate subcultural or entertainment-related material, an approach that distinguishes them from other CJK resources. For instance, \path{joujiboi/japanese-anime-speech} \cite{joujiboi2024japanese} targets automatic speech recognition in anime content, attracting high download counts and demonstrating direct utility for real-world applications such as subtitle generation. Additionally, \path{YANS-official/ogiri-test-with-references} \cite{ogiri} captures the distinctive \emph{Ogiri} comedy tradition, illustrating Japan’s unique comedy culture through multimodal data (text and images). While such resources enrich models’ ability to handle colloquial or creative contexts, they also require careful documentation of stylistic nuances and potential copyright constraints. Many subcultural datasets involve fan works or licensed content, which often preclude fully open licenses. Researchers must therefore verify these constraints to avoid unintended usage restrictions or downstream complications.

\paragraph{Diverse Methods in Dataset Processing and Refinement}
Japanese datasets exhibit a reliance on translation-based pipelines and synthetic data generation rather than building new corpora from scratch. For example, the Magpie series \cite{xu2024magpiealignmentdatasynthesis} has been adapted into multiple Japanese resources—e.g., \path{Aratako/Synthetic-JP-EN-Translation-Dataset-Magpie-Nemotron-4-20k} and \path{Aratako/Magpie-Tanuki-8B-annotated-96k} \cite{aratako_4_20k, Aratako2024}—highlighting how translations and AI-generated text can expand training data. While these strategies improve dataset availability, they raise concerns about translation errors, cultural misalignment, and potential biases introduced by synthetic text. Efforts such as \path{neody/oscar-ja-cleaned} \cite{neody_oscarja_cleaned} and \path{saillabalpaca-japanese-cleaned} \cite{upadhayay2024taco} illustrate attempts to mitigate these issues through dataset cleaning and quality control. Systematic documentation of translation processes and validation protocols would help researchers assess dataset reliability. This localization approach may serve as a model for other languages seeking rapid resource expansion.

\section{Discussion}
Our analysis reveals distinct characteristics across CJK dataset ecosystems: Chinese datasets show strong institutional backing but inconsistent documentation; Korean datasets demonstrate community-driven development but face licensing gaps; and Japanese datasets emphasize subcultural content while dealing with copyright constraints.

Three practical issues warrant attention. First, licensing diversity (particularly "Unknown" licenses in Korean datasets and restricted licenses in Japanese resources) complicates collaborative projects. More consistent adherence to guidelines like "Datasheets for Datasets" \cite{gebru2021datasheets} could enhance reusability. Second, domain clustering (medical for Chinese, subcultural for Japanese, and moderation for Korean) may underserve other areas needed for general-purpose LLM development. Third, culturally specific content requires transparent documentation, as simple translations miss nuanced cultural meanings.

Despite these differences, strong synergy potential exists. Joint benchmarks could facilitate cross-lingual comparisons, while unified documentation frameworks could standardize metadata and licensing. Our findings underscore both the richness and fragmentation of CJK resources, suggesting that clearer practices and cross-lingual collaboration can foster a robust ecosystem for East Asian LLM development.

\section{Limitations}
Our analysis primarily focused on datasets with relatively high download counts, which may have led us to overlook smaller or emerging resources that could shed light on niche trends or specialized applications. Furthermore, we limited our scope to the Hugging Face platform; investigating additional repositories (e.g., GitHub, Kaggle, or Papers with Code) could reveal a broader range of dataset characteristics and host factors. Although we manually requested permission to access certain private or restricted datasets, some ultimately remained inaccessible, thereby constraining the representativeness of our findings.

In addition, while Korean and Japanese datasets were examined with input from language experts, our review of Chinese data relied solely on documentation, potentially affecting the depth of our analysis. Finally, we chose to focus on three major East Asian languages, excluding many low-resource languages and dialects, whose inclusion could further expand and enrich our findings.

\section{Conclusion}

This study presents a comparative analysis of over 3,300 Chinese, Japanese, and Korean datasets on HuggingFace, revealing distinct ecosystem characteristics: Chinese datasets show strong institutional involvement, Korean resources are community-driven, and Japanese datasets emphasize subcultural content, highlighting that documentation, licensing, and ownership must be addressed in cultural context to guide inclusive East Asian language technologies.

\section*{Code and Resources}
All code and visualizations are available at:  

\noindent
\textbf{Korean:} \url{https://songys.github.io/Korean-HF-datasets-catalog/} \\
\textbf{Chinese:} \url{https://songys.github.io/Chinese-HF-datasets-catalog/} \\
\textbf{Japanese:} \url{https://songys.github.io/Japanese-HF-datasets-catalog/} \\
\textbf{English:} \url{https://songys.github.io/English-HF-datasets-catalog/}

\vspace{4pt}
We provide these web-based catalogs as living documents that continuously track the evolving HuggingFace ecosystem, offering researchers real-time insights into CJK and English dataset landscapes through interactive visualizations and updated statistics.

\section*{Acknowledgements}
This research was supported by Brian Impact, a non-profit organization dedicated to advancing science and technology.


\bibliography{custom}
\bibliographystyle{acl_natbib}

\end{document}